# Computational and Robotic Models of Early Language Development: A Review

Pierre-Yves Oudeyer (1) , George Kachergis (2) , and William Schueller (1,3)

1 Inria and Ensta ParisTech, France
2 Stanford University, USA
3 University of Bordeaux, France

**Abstract**: We review computational and robotics models of early language learning and development. We first explain why and how these models are used to understand better how children learn language. We argue that they provide concrete theories of language learning as a complex dynamic system, complementing traditional methods in psychology and linguistics. We review different modeling formalisms, grounded in techniques from machine learning and artificial intelligence such as Bayesian and neural network approaches. We then discuss their role in understanding several key mechanisms of language development: cross-situational statistical learning, embodiment, situated social interaction, intrinsically motivated learning, and cultural evolution. We conclude by discussing future challenges for research, including modeling of large-scale empirical data about language acquisition in real-world environments.

# 1 The role of computational and robotic models of language development[1]

Language involves a multitude of components interacting in complex ways in parallel and within several scales of time and space. There is the developmental timescale of the growth of an individual person, the glossogenetic or cultural timescale of the evolution of cultures, and the phylogenetic timescale of the evolution of species. Within the developmental time scale, ranging from moment-to-moment learning to years-long processes, multiple brain circuits interact with a multi-part body situated in a complex environment, and in various kinds of situations determining a diversity of information flows. Even though each of these levels and components needs to be studied independently in order to reduce the complexity of the problem, it is also fundamental to study their interactions. The sciences of complexity have taught us that in many natural systems, there are global phenomena that are the result of local interactions between components, but where individual study of these components would not allow us to see the global properties of the whole combined system. Many of the properties of language are probably not directly encoded by any of the components involved, but are the self-organized outcomes of the interactions of the components.

---

[1] Some material from this section was adapted from (Oudeyer, 2019, CC-BY).

**Language development as a dynamical system.** These self-organizational phenomena are often difficult to understand or to foresee intuitively, and hard to formulate in words. This is why research into language development today leverages computer science, mathematics and robotics to provide complementary methods to developmental psychology and neuroscience. These fields provide methods to build operational models of the interactions between the components involved in language, conceptualized as a complex dynamical system (Elman, 1995; Steels, 1997; Smith and Thelen, 2003; Beckner et al., 2009; Thomas et al., 2016; Samuelson et al., 2017; Oudeyer, 2018a). An operational model is one which defines the set of its assumptions explicitly and formally, and then computes or simulates the consequences of these assumptions, proving that a particular set of conclusions is entailed. Such formalization forces researchers to explicit all mechanisms and details of a theory, providing opportunities to identify gaps, as well as to do synthetic experimentation with computers and robots to test their plausibility, to understand better their implications, and to generate predictions or new hypotheses. There are several kinds of models. In particular, from the structural point of view, one can identify two broad families: analytical mathematical models, and algorithmic models of the causal mechanisms of language formation, which include a particularly important sub-family of models that consider the central role of embodiment, and that will be called here "robotic" models (as we will see, they do not necessarily entail the use of a physical robot, but may use a simulated model of the body).

**Analytical mathematical models.** A first class of models abstract a certain number of variables from the phenomenon of natural language development and express how these variables are related and how they change over time in the form of mathematical equations. Usually this comprises systems of coupled differential equations and makes use of the framework of dynamic systems theory, and sometimes statistical physics (e.g. Nowak et al., 2002; Loreto et al., 2012). When these models are sufficiently simple, their behavior can be predicted analytically and formally proved from their mathematical structure alone. However, the abstractions and assumptions on which such proofs are based are sometimes far removed from physical, cognitive and social realities. Moreover, the formal language of mathematical relations is not always very suitable for explaining processes at work in nature (or in culture).

**Algorithmic models.** A second class of models focuses on the processes of language morphogenesis: these models are formulated in terms of algorithms. These algorithms are themselves expressed, in practice, using computer programing languages. Using this kind of formal language to describe a natural or a cultural process has two solid advantages. First, the great expressivity of these languages lets them formulate highly complex processes concisely (Dowek, 2011). Secondly, in the case of phenomena whose behavior is very difficult to predict analytically from a series of equations, it can be possible to calculate this behavior automatically through simulation. The programs can be run on computers in what is a simulation of a morphogenetic process, and researchers can observe how the simulated system behaves with different parameters. In research on language development, this approach often involves building artificial systems in which individuals (their bodies, brains and behavior), their interactions and their environment are modeled by programs. A large share of models have emphasized the modeling of the processes of language representation and learning, and can be distinguished along several dimensions:

- **Learning numerical, symbolic or hybrid representations**. Some models represent language knowledge using strings or structured sets of symbols, such as for example in models of grammatical learning where sentences are represented by hierarchical trees of symbolic categories. Other models represent linguistic knowledge as sets of numbers, organized in vectors, matrices or neural activities. For examples, the discovery of invariants and associations in a multimodal flow of speech words and images can be achieved using techniques like non-negative matrix factorization (Driesen et al., 2009; Mangin et al., 2015). Another large family of models considers numerical connectionist representations, in a family of models called connectionist modeling (Westermann & Twomey, 2017). Connectionist models rely on simplified models of neural circuits and their interactions, and most often the structure of the artificial neural network is specified by the modeler and the weights of connections among artificial neurons are adapted during a simulated learning process. For example, some models of speech sound acquisition use artificial neural networks to simulate the acquisition of sensorimotor representations enabling an individual to learn to produce the speech sounds of its environment (Warlaumont et al., 2013). Some models combine connectionist and symbolic representations, such as models of various forms of syntactic processing where symbolic sequences of letters or words are fed into a neural network that learns to predict associated symbolic outputs such as word boundaries, syntactic categories or semantic parse trees (Elman, 1990; Westermann & Twomey, 2017; Dupoux, 2018).
- **Supervised, unsupervised or reinforcement learning.** Models of language acquisition leverage different types of learning processes as classified by machine learning theory. The first form of learning is supervised learning, where the model learns by observing inputs (e.g. image of an object) and associated outputs to be predicted (e.g. name of the object), trying to discover the regularities between these inputs and outputs to improve its predictions or classifications. Self-supervised learning is a specific form of supervised learning where the output is not provided by a teacher, but is directly observed in the environment by the learner (e.g. the input could be one modality of observations, and the output another modality which the learner learns to predict from the first modality). Unsupervised learning is another form of learning where inputs are not matched by target outputs, but rather the learning mechanisms attempts to discover regularities in the underlying structure of the input distribution (e.g. clustering words according to the similarity of the contexts in which they appear). Finally, reinforcement learning is a form of learning where the learner attempts to learn a (context-dependent) behavioral policy that maximizes a reward expressed in the form of a scalar feedback. Hence, the learner is not provided examples of correct behaviors, but rather its behavioral attempts receive a score of fitness that it tries to improve. For example, some models of speech learning use an intrinsic reward measuring the saliency of sounds to drive the learning of vocalizations (Warlaumont, 2013), and some other models use extrinsic reward measuring task completion to drive the acquisition of sentence interpretation in joint task contexts (Daubigney et al., 2012).
- **Normative or heuristic models of learning processes**. Another dimension of importance in the landscape of models of learning and representational mechanisms is the normative/heuristic distinction. One approach that has been very fruitful has been to model the child language learner as a Bayesian and/or rational learner which optimally infers structures from observed data and from a priori knowledge expressed in the form of probabilistic priors (Chater & Manning, 2006; Tenenbaum et al., 2011). The advantage of such an approach, especially when expressed in the Bayesian framework, is that it forces to express formally and in a simple mathematical language most of the model assumptions, and it provides principled mathematical tools to evaluate the goodness of fit of particular models to account for a body of empirical data. However, such an approach has also several drawbacks. First, a number of findings show that the human brain, and especially the child's brain, may often learn in non-optimal and non-Bayesian manners, using heuristic inference, cognitive shortcuts, not using all information available, and prone to various forms of errors (Morevedge & Kahneman, 2010). There are actually a number of arguments showing the potential evolutionary selection of such heuristics in rapidly changing environment with severe limits of cognitive and metabolic resources (Todd & Gigenrenzer, 2000; Oudeyer, 2018b). Another drawback of normative Bayesian models is that they require pre-specification of all possible observations, events and models in order to be able to compute probabilities. Hence, by construction they do not address the question of how representations are learned (they only address the question of how certain representations are selected among existing ones), which is a fundamental question of language development (e.g. how do phonetic representations form? How are word meaning representations formed? How are syntactic categories formed? etc). Also, normative Bayesian models become computationally intractable if one aims to scale to real world high-dimensional data (Bossaerts & Murawksi, 2017). For these reasons, another very large family of models relies on heuristic models of learning, ranging from connectionist approaches (Westermann and Mareschal, 2014, Cangelosi and Schlesinger, 2015; Twomey et al., 2016) to heuristic statistical learning (Mangin et al., 2015) or symbolic learning (Mealier et al., 2017; Spranger and Steels, 2012). The advantage of these models is their very large expressivity, their capacity to address the problem of representation learning (especially in connectionist approaches), and their capacity to combine different kinds of learning mechanisms in the same model. Also, these models have shown to scale better to complex real world situations, as shown by their pervasive use in robotic models of language learning grounded in high-dimensional spaces of perception and action (Cangelosi et al., 2010). A drawback of these models is that they often contain many free parameters, and there are no principled unified statistical framework enabling to compare in an unequivocal manner their goodness of fit to account for empirical data.

Beyond a view of the computational modeling landscape organized along the formal technical dimensions we just described, there are two other ways to structure this landscape. First, it is possible to classify models in terms of which stage(s) of infant language development they are focusing on. In this chapter, we are focusing on the models of early language development, ranging from vocal development to the onset of speech as a communication medium and to early word learning. Many other models in the literature have also focused on later stages of language development, with a large focus on the development of syntactic capabilities: for these works, we refer the readers to these excellent reviews: Chater and Manning, 2006; Monaghan and Christiansen, 2008; Yang, 2011; McCauley and Christiansen, 2014.

Another way to classify models is in terms of the general causal developmental mechanisms that they focus on. This is the approach we follow in this chapter, where we analyze models in terms of several of these general causal mechanisms: cross-situational statistical learning, embodiment, social interaction, self-organization of brain-body-environment couplings, intrinsic motivation, and the links between learning and evolution.

Many computational models of language learning focus largely on the learning mechanisms involved in mapping words to their intended referents, referred as the problem of cross-situational learning (see section 2): the mechanisms used to detect regularities in language data, simplifying models of the interaction with the environment, of how data is collected, and how this impacts the properties of data. For example, many works model the environment as a database of examples which are incrementally and randomly selected by the learner to train their learning mechanism (e.g. a database associating words with their potential meanings). We focus on models for learning meaning, but many of the issues we highlight are relevant for models of syntax learning as well.

**Robotic models: embodiment, social interaction and intrinsically motivated active learning.** However, as we will detail more in sections below, the real world environment, and the way it is perceived and acted upon by an active body, and through situated interaction with others, contains a lot of structure that can guide learning processes. This is why computational models of language development have in recent years been hybridized in implementations that combine the use of computers and robotics. These models, instead of representing the brain-body-environment system purely as computer algorithms, make use of programs running on computers that are embedded in physical robots. Here, only the "brain" is formulated algorithmically, while the body is modeled using mechatronic elements, and the environment approximates to humans' real environment (including interaction with social peers though human-robot interaction). Furthermore, the "brain" in these approaches is not simply viewed as a passive statistical learning system, but rather an intrinsically motivated and goal-directed system. This approach is currently at the heart of developmental robotics (Cangelosi and Schlesinger, 2015), where there is intense activity around the modeling of cognitive sensorimotor, cognitive and social development (Oudeyer, 2010). Finally, beyond the body and motivational context, language development has also been analyzed from the perspective of its bi-directional interaction with cultural and phylogenetic evolution (Steels, 2003; Cangelosi et al., 2010).

**Symbol grounding problem**. Building this kind of robotic model is interesting from several points of view. First, regarding language, it addresses the symbol grounding problem, in other words the fundamental problem of how symbols are grounded in the physical world. The problem here, so eloquently formulated by Steven Harnad (Harnard, 1990) is understanding how the symbols commonly used for describing and modeling languages (such as words and grammatical rules) can become meaningful in the physical and social reality of a real organism. In particular, it involves a capacity to link the abstract world of symbols to the concrete world of numerical and chemical quantities that are perceived and manipulated by the brain and the body in context. It is hard to see how models that are purely algorithmic, implemented entirely on computers that at base represent the world in a symbolic, discrete way, might usefully inform any questions about symbol grounding. Hybrid algorithmic-robotic models, on the other hand, that by definition are at the edge between the symbolic and the physical worlds, are extraordinary tools for studying this problem (Steels et Kaplan, 2001; Steels, 2012). Here algorithmic models of language are confronted with physical and social reality, and whether or not there is an effective grounding of symbols – a strong constraint on the plausibility of these models – can be tested empirically.

**Brain-body-environment dynamic interactions.** There is another very good reason for using robots. As Esther Thelen and Linda Smith, for example, have argued in their theory of development (Smith and Thelen, 2013), the formation of behavioral and cognitive structures results from dynamic interaction between the brain, the body and the environment. The body and the environment, whose physical substrate gives rise to particular properties of structure generation, have a crucial role. Embodiment, that is to say the material composition and the geometry of a body and its sensorimotor system, can dramatically simplify the acquisition of certain behaviors. For example, Chen Yu and Linda Smith showed how geometrical hand-eye relations and the physical manipulation of objects could create favorable situations for learning the meanings of first words (Yu and Smith, 2012). Thus, the body carries out physically a type of information processing, sometimes referred to as "morphological computation" (Pfeifer et al., 2007). In this context, robots make it possible to model mechatronically – in a straightforward, realistic way – interactions between the brain, the body and the environment that would be far too complex or even impossible to model algorithmically. The section on the role of embodiment below provides several examples of robotic models studying this perspective. In other spheres, many other examples can be found today of robotic models being used to enhance understanding of animal and human behavior (Oudeyer, 2010), concerning such varied phenomena as navigation and phototropy in insects, control of locomotion in dolphins, distinguishing between self and non-self in human infants, but also the impact of the visual system on the formation of linguistic concepts.

# 2 Mechanisms of Cross-situational Learning

Beyond learning the sounds of a language (e.g., Feldman, Griffiths, Goldwater, Morgan, 2013), and segmenting contiguous speech into words (e.g., Monaghan & Christiansen, 2010), infants face the daunting challenge of mapping words to referents. Each utterance from a caregiver might have any of a variety of intended meanings that can be difficult to discern in spite of cues such as gaze and pointing. A useful source of disambiguating information may be had if infants are able to track which words are frequently heard in conjunction with particular referents. If a given word and its intended referent co-occur repeatedly across a variety of scenes, and infants can track this conjunction to some extent, then the word's meaning may be learned cross-situationally. Such cross-situational learning is thought to be an important way for children to learn words with concrete referents (Smith & Yu, 2008; Pinker, 1989). Since most real-world situations contain many possible word-referent mappings, but time and attention are limited, learners likely use heuristics or strategies (implicit or explicit; learned or innate) to restrict the number of meanings they consider. Computational models of cross-situational learning have sought to discover the representations and mechanisms that people use to track and disambiguate word-referent co-occurrences.

Two dominant modeling approaches have emerged to account for cross-situational word learning: associative models and hypothesis-testing models. These accounts are discussed below with details of how they are typically compared to human learning behavior in short-term experiments. However, models of both types often make simplifying assumptions about the input (e.g., pre-segmented words and well-defined objects) that are unlikely to be met in the real-world language environment. Models proposed in developmental robotics use machine learning models with fewer assumptions about the input, but also have not yet been shown to scale to real-world input. Independently, a variety of simulation investigations have asked whether simple learning mechanisms can scale to acquiring a full-sized adult lexicon in a reasonable amount of time. For analytical purposes, these simulations often make unrealistic simplifying assumptions about the independence of learning each word, and about the sampling process by which words and referents are experienced.

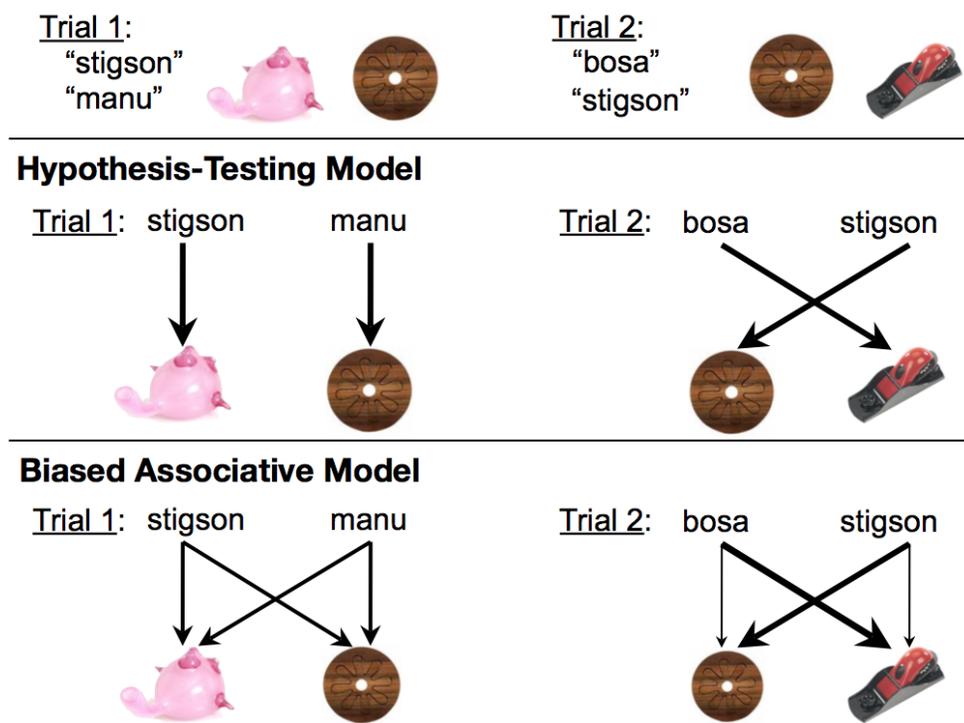

*Figure 1: Example of two cross-situational learning trials (top) from which three word-referent pairs might be learned, along with schematic representations that might be learned by a hypothesis-testing model (middle; e.g., propose-but-verify) that tracks a single hypothesized referent per word, and for associative models (bottom; e.g., the familiarity- and uncertainty-biased associative model) that attends to all co-occurring words and objects to some extent.*

Cross-situational word learning has been studied in infants (Smith & Yu, 2008), children (Akhtar & Montague, 1999), and adults. While it is assumed that many of the same word-learning mechanisms are operating in adults, it is important to note that cognitive abilities in infants and children are still developing. Adults are often studied because their greater attention span enables the use of more complex, extended designs. In a typical adult study, participants are asked to learn the referent of roughly twenty nonce words by watching a series of training trials. On each trial, learners see an array of two to four unfamiliar objects (e.g., sculptures) and hear a corresponding number of pseudowords (e.g., *stigson, manu*). See Figure 1 (top) for an example of two cross-situational learning trials, from which an observer may learn three word-object mappings. Although each word refers to a single onscreen object, the referent of each pseudoword is ambiguous on a given trial, because the intended referent is not indicated. In a typical learning scenario, participants might view each of 18 word-object pairings six times as they appear four at a time across 27 trials, for a total duration of several minutes (Yu & Smith, 2007; Kachergis, Yu, & Shiffrin, 2013; Suanda & Namy, 2012).

**Hypothesis Testing Models.** The hypothesis-testing theories view word learning as a problem of induction with an enormous hypothesis space that must be reduced by the learner applying a number of language-specific constraints to in order to simplify the problem (Markman, 1992). In this view, infants generate hypotheses that are consistent with this set of constraints and principles. For example, the global principle (or bias) of mutual exclusivity (ME) assumes that every object has only one name (Markman & Wachtel, 1988). At a lower level, the fill-the-lexical gap bias is proposed to cause children to want to find a name for an object with no known name (Clark, 1987; Merriman & Bowman, 1989). When given a set of familiar and unfamiliar objects, it has been shown that 28-month-olds assume that a new label maps to an unfamiliar object (Mervis & Bertrand, 1994). Similarly, the principle of contrast states that an infant given a new word will seek to attach it to an unlabeled object (Clark, 1987). Fill-the-gap, ME, and contrast make many of the same predictions made by the more general novel name-nameless category principle (N3C), which states that novel labels map to novel objects (Golinkoff, Mervis, & Hirsh-Pasek, 1994). In order to be of aid to infant learners, such principles are thought to be either innate or developed very early in life (Markman, 1992).

The hypothesis-testing approach is used in the formal analysis of language acquisition (Gold, 1967; Pinker, 1979), stemming from inferential methods in the philosophy of science, with developmental theories built upon this rationale (Carey, 1978; Clark, 1987). A shared intuition among these approaches is that the multitude of co-occurrences available in the visual and auditory environment of the learner is far too complex to be tracked, stored, and updated (Medina, Snedeker, Trueswell, & Gleitman, 2011; Trueswell, Medina, Hafri, & Gleitman, 2013). A model that exemplifies the hypothesis-testing approach is the propose-but-verify model (Trueswell et al., 2013), which assumes that only a single hypothesized referent is stored for each word, and that on a further exposure to the word this proposal is recalled with some probability, and then either verified if the referent is present–increasing the future probability of recall, or discarded if the referent is absent (see Figure 1, middle). In case the hypothesized referent is absent or fails to be retrieved, a new hypothesis is chosen from the currently available referents. This model accounts for adult word learning behavior in some experimental contexts (Trueswell et al., 2013), but not in others (Kachergis & Yu, 2017). More sophisticated rule-based models of word learning scale well in some corpus-based applications (Siskind, 1996), but have not attempted to account for human behavior in experiments, or in real-world learning environments. A Bayesian model of cross-situational word learning makes binary word-referent hypotheses according to the global co-occurrence structure, combined with a prior preferring a small lexicon (Frank, Goodman, & Tenenbaum, 2009). This model is able to learn small lexicons from child-parent interactions, based on transcribed speech and hand-coded representations of the visible objects. These models showcase the learning power of sparse, hypothesis-based representations in combination with rules and biases of which hypotheses to form.

**Associative Learning Models.** In another view, word learning can be explained as a gradual accumulation in memory of all experienced co-occurring words and referents, perhaps relying on domain-general associative mechanisms rather than positing logic-based constraints or language-specific mechanisms (Smith, 2000; Regier, 2005; Kachergis, 2012). In general, these models accumulate associative strength between any co-occurring word and object, but the increment of a particular association can be weighted by factors such as prior knowledge (i.e., an already-existing association based on past co-occurrence), novelty, or uncertainty (the entropy of a word or referent's associations; see Figure 1, bottom). This online attentional shifting can allow associative models to show order effects such as highlighting (Kachergis, 2012), bootstrapping of low frequency word meanings when they appear in the context of known referents (Kachergis, Yu, & Shiffrin, 2016), and inference-like behavior much like a mutual exclusivity bias (Kachergis, Yu, & Shiffrin, 2012). These behaviors of the biased associative model emerge from the interaction of competing attentional biases to 1) strengthen already-strong associations, and 2) attend to words or referents uncertain (high-entropy) associations (Kachergis et al., 2012). Rather than using arbitrary associative strengths, some associative models maintain conditional probability distributions that are updated as new co-occurrences are observed (Fazly, Alishahi, & Stevenson, 2010). The intuition behind associative learning models is that any words heard in a given context will impact the representation of the referents–to some extent. While it may seem too difficult for a learner to track the associations between a word and not only its intended referent, but its many distractors, the presence of the unintended associations both provides a sense of context (e.g., forks and knives often appear together), and serve as noise when learners are trying to retrieve the correct association. Associative models tend to be able to account for detailed behavioral effects found in experiments, including interactions of context diversity and word frequency (Kachergis et al., 2016) as well as response trajectories during word learning (Kachergis & Yu, 2017). Recent efforts to match detailed human learning trajectories across a range of experimental conditions have found that sampling versions of models—both Bayesian (Yurovsky & Frank, 2015) and associative (Kachergis & Yu, 2017)--best match human behavior by storing multiple (but not all possible) hypothesized referents for each word. It has been pointed out that simple hypothesis-testing (e.g., Medina et al., 2011) and simple associative accounts are at the endpoints of a continuum of sampling models (Yu & Smith, 2012). A growing family of models combine associative learning with online referent selection to achieve better fits to empirical data (e.g. McMurray, Horst & Samuelson, 2012; Kachergis & Yu, 2017).

**Developmental Robotics Models.** However, as mentioned earlier, both hypothesis-testing and associative models of cross-situational word learning usually operate on the pre-segmented words and objects that are assumed to be easily identified by adult learners in cross-situational learning studies (Yu & Smith, 2007, Kachergis et al., 2016, Suanda & Namy, 2012). Modeling these short-term experiments has yielded important insights into how people store and update particular word-referent associations (or hypotheses), but using pre-segmented words and objects that are distinct from each other and from the background may oversimplify the mapping problem. Real-world learning environments offer a stream of changing multimodal information, from which learners must extract regularities at the appropriate level. It may be that learners need to first learn phonemes, then words, and even word classes before being able to map the appropriate entities to referents. Similarly, learners may need to learn to segment the visual world into objects, properties, and actions before being able to map words to these entities. However, learners may be able to take advantage of regularities in the cross-modal representations of both auditory and visual information (and other modalities, even). This is the approach taken by developmental robotics researchers, who have investigated cross-situational word learning using machine learning algorithms that build shared, cross-modal representations of scenes and utterances, from which words and concepts can be extracted (Mangin, Filliat, ten Bosch, & Oudeyer, 2015; Chen et al., 2018). Such models suggest how it is possible to simultaneously learn different types of words, referents, and their associations without assuming separate processes for learning the structures within each modality (e.g., object detection or word segmentation). Nonetheless, these models are often tested in artificial experiments with small numbers of objects, utterances, and actions, and will require more effort to be scaled up to realistic situations and to be compared in detail to human word learners.

**Large-scale Simulations.** Efforts to understand whether cross-situational learning can realistically learn an adult-sized vocabulary come from simulation studies investigating simple learning mechanisms. (Blythe, Smith, & Smith, 2010) tested how quickly a logic-based fast-mapping mechanism (that strictly rules out any referents not currently present when a word appears–often resulting in 1-shot learning) can be expected to learn the 60,000 words in an adult-sized vocabulary. The simulation showed that learning time for a full vocabulary using a fast-mapping mechanism is well within reason, with 99% of the words learned by the time 940,000 words have been sampled (i.e., 142 words per day for 18 years). Simulations of a hypothesis-testing "guess-and-test" mechanism had learning times that were only 50% slower–requiring a still reasonable 214 learning episodes per day. However, these analytical estimates rely on making a variety of simplifying assumptions that likely impact the validity of these estimates. The assumptions made by Blythe et al. (2010) and others (Blythe, Smith, & Smith, 2016; Vogt, 2012) are: 1) a word is only heard when its meaning is present in the situation, 2) perception of words and situations is errorless, 3) every situation has the same number of possible referents, 4) each word maps to a single meaning, 5) learners know the space of meanings that are possible, and 6) words are assumed to be learned independently, meaning learners are expected not be using even a mutual exclusivity bias. Many of these assumptions further simplify the learning problem, although critically, some are not realistically plausible (e.g. words are often heard in the absence of the referent). Furthermore, the distributions of words, referents, and situations are independently randomly sampled in these simulations, rather than reflecting the skewed frequency distributions found in real-world speech and the nested structure found in natural scenes (Hidaka, Torii, & Kachergis, 2017). Future studies will need to consider long-term learning in simulations using more realistic learning mechanisms, as well as more realistic distributions of experience, with interdependent word learning.

**The statistics of the language environment**. The exponential distribution of word frequency–with a few words appearing quite often, and the majority of words being vanishingly rare–has intrigued researchers for decades, but there is no consensus on the cause of it (Zipf, 1949; Piantadosi, 2014). The fact that many words are infrequently heard presents a difficulty for learning, as there will be few opportunities to learn these words and disambiguate their meanings. If referents also have an exponential frequency distribution, simulations predict that cross-situational learning will be orders of magnitude slower (Vogt, 2012). Much recent research has been focused on the distribution and structure of both the language input that children receive (Jones & Rowland, 2017; Hart & Risley, 1995), as well as the visual scenes they encounter. While we often think of visual scenes as offering a multitude of possible referents (Medina et al., 2011), analysis of head-mounted cameras on infants during free play has shown that infants often have only a single object dominating their view (Smith, Yu, & Pereira, 2011). The object filling their visual field is often being actively manipulated by the child or caregiver, and is likely to be mentioned. This finding seems to deflate the fabled complexity of the cross-situational learning problem. Importantly, the studies reviewed so far primarily address learning the semantics of words, largely ignoring syntax. However, a variety of models have been proposed to address the problem of learning syntax from experiencing sentences (e.g., Chang, Dell, & Bock, 2006; Freudenthal et al., 2007; Thomas and Knowland, 2014), as well as for simultaneously learning semantics and syntax (e.g., a Bayesian model: Abend et al., 2017; a connectionist model: Li, Farkas, & MacWhinney, 2004). Due to space limitations we will not cover syntax learning in-depth, but many of the issues we identify for cross-situational word learning models also apply to syntax learning models.

Rather than making assumptions about the learning process, some researchers have opted to use statistical waiting time models to investigate the impact of factors such as word length, word class, frequency in child-directed speech, and imageability to predict words' age of acquisition (Braginsky, Yurovsky, Marchman, & Frank, 2016; Hidaka, 2013; Mollica & Piantadosi, 2017). Fitted to children's vocabulary growth curves, these statistical models have been used to estimate the number of exposures and whether the rate of learning changes during development (Hidaka, 2013; Mollica & Piantadosi, 2017). Another approach is to use network theoretic models and semantic relatedness measures to model the growth process of children's vocabulary (Hills, Maouene, Riordan, & Smith, 2010). Much research has been devoted to characterizing individual variability in early word learning, finding that the amount of child-directed speech children receive correlates with vocabulary size and school readiness (Hart & Risley, 1995; Huttenlocher, Haight, Bryk, Seltzeer, & Lyons, 1991). Recent efforts to create large shared databases of early word learning data such as WordBank (Frank, Braginsky, Yurovsky, & Marchman, 2016) and recordings of child-directed speech in the home such as HomeBank (VanDam et al., 2016) and CHILDES (MacWhinney, 2000) promise to unveil more about the structure and content of children's language input and its effects on vocabulary learning, and to serve as new goalposts and constraints for modeling efforts.

However, children do not learn only by passively receiving this audiovisual stream of information. Rather, they actively use their body to explore language and its referents, and they are active communicative partners, likely attending to and leveraging a variety of social cues to aid their learning. The next sections review how computational models have approached these active dimensions of language development.

# 3 Robotic models: The role of embodiment, situatedness and social interaction

Children language learners are distinguished from most artificial machine learning systems in that they are situated in a real-world physical environment with social peers. They are embodied in a very complex physical organism which filters information from the environment and affords certain kinds of actions on the environment. Also, children learn language in the context of achieving social and material goals. This situatedness and embodiment provides multiple forms of constraints that both define and guide language learning processes. Robotic models have been developed to study the roles and structures of these constraints. It is important to note that while robotic models focus on the physical and situatedness properties of the learner, they are not necessarily implemented with real world physical robots. Indeed, many robotic models are entirely implemented in a virtual world with a simulator of the robot body and its physical/social interaction with the environment. The advantage of using simulated robotic bodies is that it facilitates systematic experimentation (gathering more statistics, larger exploration of the space of parameters). The drawback of simulated robotic models is that some aspects of the real world are difficult to simulate adequately, such as the natural variations (structured noise) in perceptual channels or social interaction with a (model) caretaker. Let us now give a few representative examples of (simulated or real world) robotic models of language development.

**Leveraging spontaneous structure from physics and body morphology.** Models of speech development have extensively studied the role of physical embodiment, due to the central role of the vocal tract and auditory system in phonetic and phonological learning (Stevens, 1972; Schwartz et al., 1997). The vocal tract is one of the most complex organs in the body, where a large number of muscles are used to continuously deform soft material parts such as the tongue, lips or vocal folds (Boersma, 1998). In addition, the produced sounds are perceived by a complex auditory system with many non-linearities (Schwartz et al., 1997). One scientific challenge in speech learning is understanding how children learn to produce the speech sounds of their native language given the high complexity of this sensorimotor system. From the point of view of control theory, this appears to be a conundrum given the high-dimensionality of the space and severe limits on time and energy available to the child for trying out vocal tract movements (Bernstein, 1967). So how can children learn canonical speech sounds already by the end of their first year? Several models have studied the natural dynamics of vocal tract movements, resulting from both mechanical coupling of movements and neural synergies among articulators. For example, Kelso (Kelso et al., 1986) took a dynamical systems approach, showing that random motor commands sent to the vocal tract produced already highly structured movements (hence speech sounds) due to the spontaneous structure resulting from these coupling dynamics. This enables to show how learning speech sounds may amount to tuning some parameters of these spontaneous structures, which is much easier than learning from scratch and without constraints high-dimensional movements of the articulators.

**Early words are grounded in concrete action and perceptual repertoires.** The role of embodiment has also been emphasized in several models of early word learning. A first obvious reason is that the first words children learn are concrete and directly related to their bodies, their actions and their interactions with the environment (Bloom, 1995). Thus, the representation of the meanings of these first words, especially verbs and nouns, is intrinsically defined in terms of the action repertoire of children, and the way they perceive visually, haptically, auditorily or spatially the objects around them. From this perspective, robotic models of bodies and their physical perception and action on the environment are a prerequisite for modeling the acquisition of the meaning of these early words, such as shown in models of learning the names of shapes and colors (Steels, 2001; Roy, 2005), simple manipulative actions (Cangelosi et al., 2010) or spatial relationships among objects (Spranger & Steels, 2012).

**Social scaffolding and imitation.** A second straightforward role of embodiment and situatedness, studied in robotic models, relates to social interaction and the non-verbal cues in the gaze, gestures, and discourse used by the language partners. These non-verbal cues enable joint attention, and more generally social guidance. Infants have been shown to attend to a variety of social and discourse cues, which may greatly simplify both speech segmentation and word learning. For example, the speech directed at infants by caregivers (e.g., *motherese*: child-/infant-directed speech) is characterized by intonation and prosodic cues, and has been shown to aid segmentation (Thiessen, Hill, & Saffran, 2005). It has been shown that adults are better able to segment artificial languages when given prosodic cues, which seem to serve as a filter on the learned transitional probabilities (Shukla, Nespor, & Mehler, 2007).

Beyond speech segmentation, social cues can reasonably be expected to impact word learning as well. Experiments have found that infants can follow a speaker's gaze to infer what they are referring to (Baldwin, 1993) Episodes of joint attention between an infant and caregiver, as when playing with toys together, show characteristics such as shorter sentences from the caregiver and more utterances from both in the dyad (Tomasello, 1988). Caregivers referring to objects that were already focused on by the infant were correlated with a larger vocabulary, while children of caregivers who attempted to redirect their attention had smaller vocabularies.

Other research seeking to characterize natural interactions between caregivers and children has found that child-directed speech is quite repetitive, with repetitions of phrases, not just single words (Snow, 1972). An analysis of parent-child interactions while playing with toys showed the informativeness of a variety of social cues relating to the hands and eyes of the speaker, as well as to the continuity of discourse about particular referents (Frank, Tenenbaum, & Fernald, 2013). No single cue served as a perfect filter for the cross-situational learning of words, but in combination these cues much reduce the ambiguity of intended meanings. Hearing an utterance, infants may jointly consider the uncertainty about a speaker's intended meaning as well as uncertainty about the meaning of each word. This framing of the problem as one of communicative inference is the basis for a model that simultaneously learns intended word-referent mappings as well as the relative value of social cues in making such inferences (Johnson, Demuth, & Frank, 2012).

Early robotic models of language acquisition (Steels & Kaplan, 2000) compared the quality of learning input (level of ambiguity between utterances and perceived scene) provided to a robot learner in situations where 1) the human is socially and physical engaged in the interaction, synchronizing pointing gestures towards referents while monitoring the gaze of the robot to ensure gestures and referents are attended at the right moments; 2) the human is semi-engaged, only using utterances but not using actions to drive the learner's attention; 3) the human is not socially and physically engaged, only describing the scene with utterances, independently of what the robot is currently looking at. This kind of model allows us to quantify the additional learning efficiency resulting from these various levels of engagement, leveraging embodiment and situatedness. Other works designed specifically robot learners capable to move not only to act upon objects, but to communicate with social peers and realize joint attention (Scassellati, 1999). Some recent lines of work have used robotic models of embodied social language learners to also study how humans naturally teach language, how they use social cues to provide feedback, for example using motherese (infant directed speech) or *motionese* to demonstrate simplified and highly informative learning examples (Vollmer & Schillingmann, 2017).

Developmental robotics models have further studied various links between sensorimotor learning and social language learning. For example, a model based on intrinsically-motivated learning for efficient coding via active perception learns to copy goals, rather than the specific motor movement, allowing it to learn simple behaviors such as gaze-following (Triesch, 2013). The model begins by observing a tutor's behavior and models the sensory consequences of the behavior. Next, the model acts and receives a reinforcement signal from within that encodes how well its sensations are matched by the sensory model. The model's behavior is adapted to make the sensory consequences of its actions better match sensory model learned from watching the tutor's actions.

Cederborg and Oudeyer (2013) introduced a model for learning to acquire multiple skills by observing a tutor's ambiguous demonstrations. The model integrates concepts and techniques from earlier cross-situational learning models, as well as models of motor learning by demonstration that treat meanings as complex sensorimotor policies with coordinate systems that must be inferred. A contribution of Cederborg and Oudeyer is that the model learns both linguistic and non-linguistic skills in a single process, without specifying a linguistic channel to the model. The proof-of-concept demonstrates the viability of this approach, and future investigations will be needed to determine how well it scales, and how well it matches human developmental trajectories.

**Multimodal regularities.** Other models have studied more subtle, but equally fundamental, roles of embodiment. In addition to a flow of passively perceived utterances and visual scenes, embodiment and situatedness provide the learner with the opportunity to also observe concurrently a flow of actions and effects on the scene (including proprioception). This additional flow of information, enabled by embodiment and consisting in an action-oriented modality, contains structure which can often facilitate statistical inference of ambiguous structures and associations in the linguistic domain. For example, Mangin et al. (2015) show how invariants (e.g. words) in low-level unsegmented speech streams, as well as their combinatorial structure and associations with objects and actions, can be learned jointly with invariants and structure in low-level flows of images and action movements using multi-modal cross-situational learning methods. Like in other related models (e.g. Cangelosi et al. 2016; Sugita & Tani, 2005; Mohammad et al., 2009), such correlated flows of linguistic and sensorimotor information enables inference of general structures of sentences and generalize, i.e. understand the meaning of new sentences which precise word sequence was not encountered during training. Another example of the facilitating role of sensorimotor information flows in language learning is the embodied model of linguistic number counting presented in De la Cruz and colleagues (2014). Here, a neural network model is used to account for how children might learn to count linguistically by pronouncing the numbers in sequence, and how this might bootstrap internal representations of numbers that link the names of numbers to a meaningful underlying number representations. The model compares a situation where the neural network is only observing sequences of linguistic names, and a situation where the network is also concurrently observing proprioceptive information of finger counting actions: experiments have shown that observing proprioceptive finger information enables to improve both the accuracy of counting and the quality of the acquired internal representations of numbers. Interestingly, the mediating effect of sensorimotor representations for language learning has also been used to model surprising effects of posture during word learning experiments (Morse et al., 2015), reproducing observations of Samuelson et al. (2011) that the inference of word meanings referring to objects can be significantly influenced by the posture they have when children hear these novel words.

# 4 Curiosity, intrinsic motivation and active learning in language development

Learning in general, and language learning in particular, is not simply a passive information-processing cognitive phenomenon in children. It is also a motivated and active process of enquiry of the external world. Indeed, to learn speech and language, children spontaneously explore how their vocal tract can produce a variety of sounds, how these sounds produce social effects in their peers, and systematically point to objects or ask questions about all kinds of things to get linguistic information. Doing so, they collect large databases of learning examples that are needed to learn language (Oudeyer & Smith, 2016), but this is very costly in brain time and energy: exploration, trial and errors, conversations take a lot of time; and processing the initially unknown linguistic data requires significant cognitive effort. This is particularly mysterious in early speech development, when infants have not yet understood what language is, or how it can be used as a tool to fulfill external goals (e.g. asking a caretaker to bring an object). Nonetheless, infants systematically spend a lot of energy exploring what sounds they can make with their vocal tract (Oller, 2000), not knowing that it may be useful for the yet-to-be-discovered language tool. So one fundamental question is: why do they spend so much energy exploring and learning language? What are the proximal mechanisms that push them to be interested in language learning at the developmental level? Several theoretical perspectives can be taken (Oudeyer, 2006). One consists of speculating that evolution might have selected language-specific motivational circuits that push infants to explore language. However, no precise naturalized formulation of this approach has been articulated so far, in particular in terms of how these language-specific motivational circuits might be developmentally implemented and how they can address the variety of exploratory language activities. Another perspective is that of social learning, and imitation in particular. While imitation learning indeed plays a major role in language learning (Bloom, Hood, & Lightbown, 1974; Kuhl, 2000), and has been the focus of several computational models (Cederborg & Oudeyer, 2013), it cannot easily account either for the full range of spontaneous language exploration shown by infants: adult speakers do not themselves show such systematic exploration (e.g. they do not babble spontaneously all kind of sounds), and thus imitating the motivated exploration process is not a possibility. Also, while adults provide feedback and encouragement, the general difficulty of using this mechanism to drive the interest of children towards various kinds of activities is not fully compatible with the universal and systematic spontaneous interest shown by children in language exploration.

**Early language acquisition driven by curiosity.** Another hypothesis is that children are equipped with general intrinsic motivation systems that push them to explore their body, their actions, and their vocal tract and language, through curiosity-driven learning (Oudeyer and Kaplan, 2006; see further review in Chapter 6, Twomey and Westermann, this volume). This hypothesis has been fleshed out in the last decade through a series of computational models of curiosity-driven learning in developmental robotics (Oudeyer and Kaplan, 2006; Oudeyer and Smith, 2016; Forestier and Oudeyer, 2017). Considering child development in general, psychologists have proposed early on that humans and other animals may be equipped with intrinsic motivation neural circuits that push them to explore activities and stimuli for their own sake (Berlyne, 1960), as opposed to maximizing an external reward such as food or social feedback. These motivational neural circuits were proposed to rely on intrinsic rewards measuring interestingness of stimuli or activities in terms of quantities such as novelty, surprise, cognitive dissonance or intermediate complexity (Berlyne, 1960; White, 1959; Festinger, 1957). Within the last decade, research on computational modelling of these processes of spontaneous exploration driven by forms of curiosity have provided several formal frameworks used to understand these mechanisms (Gottlieb et al., 2013; Friston et al., 2017; Oudeyer, 2018), and to experiment in robotic models of sensorimotor learning (Baldassarre and Mirolli, 2013). Within this line of research, some models have considered the hypothesis that intrinsic rewards in humans are provided by sensorimotor activities which provide progress in learning predictive models of the world (Kaplan and Oudeyer, 2007), and showed how it could enable robot learners to spontaneously discover skills of increasing complexity (Oudeyer et al., 2007; Oudeyer and Smith, 2016). In the Playground Experiment (Oudeyer and Kaplan, 2006), the sensorimotor space included both leg/head movements and (simulated) vocal tract movements, as well as perception of object movements, visual saliency and sounds. The environment contained both physical objects with affordances, as well as an "adult" peer robot contingently imitating the vocalization of the learning robot. This experiment showed that the same general curiosity-driven learning system leads the robot learner to orderly explore and learn how to act upon objects with its legs/head, and how to provoke vocal responses from the peer robot by producing vocalizations, i.e. to enter spontaneously in a primitive form of speech interaction.

Using a similar model, but focusing on the study of the explored vocal sounds, Moulin-Frier et al. (2014) conducted experiments where a robot explored the control of a realistic model of the vocal tract in interaction with vocal peers through a drive to maximize learning progress. This model relied on a physical model of the vocal tract, its motor control and the auditory system. The experiments showed how such a mechanism can explain the adaptive transition from vocal self-exploration with little influence from the speech environment, to a later stage where vocal exploration becomes influenced by vocalizations of peers. Within the initial self-exploration phase, a sequence of vocal production stages self-organizes, and shares properties with infant data (Oller, 2000): the vocal learner first discovers how to control phonation, then vocal variations of unarticulated sounds, and finally articulated proto-syllables. As the vocal learner becomes more proficient at producing complex sounds, the imitating vocalizations of the teacher provide high learning progress resulting in the well-known infant shift from vocal self-exploration to vocal imitation (Oller, 2000).

**Discovering the linguistic function of speech utterances.** More recently, Forestier and Oudeyer (2017) extended these models in experiments showing not only how curiosity-driven exploration could lead a learner to explore its vocal tract, but also how it could lead to learning how to use these speech sounds to manipulate simulated social peers, e.g. getting them to bring an interesting object which is out of physical reach. Furthermore, the developmental trajectories generated in these simulations share several qualitative properties with infant development, e.g. overlapping waves phenomena in tool use development (Siegler, 1996), showing the plausibility of the hypothesis that curiosity may play a crucial role in the onset of language development.

**Active learning generates an ordered curriculum for learning.** Finally, this family of models also enables to study the links between curiosity and the impact of curriculum learning in language acquisition. As reviewed in Smith et al. (2018), the input statistics children are exposed to during their first years evolve with time along a learning curriculum that controls the growth of complexity in the perceived situations. This has been argued to be key in enabling children to learn efficiently and quickly the concepts that are needed for language grounding. The actions and attentional decisions made by infants play a key role in structuring this input, and models of curiosity-driven learning have shown how mechanisms searching for niches of learning progress push the learner to focus on activities or stimuli of gradually increasing complexity (Kaplan and Oudeyer, 2007). In other words, curiosity can be viewed as a mechanism for actively controlling the growth of complexity in language formation (Schueller and Oudeyer, 2016). Indeed, an empirical study of adult learners shows that giving them active control over which objects they will see named on the next trial significantly improves their word-learning performance, as compared to passively viewing a randomly-ordered selection (Kachergis et al., 2013). In simulations of learning a full-sized vocabulary with Zipf-distributed word and referent frequency distributions, such self-directed control over the to-be-named referents was necessary in order to make learning possible on a realistic timescale (Hidaka et al., 2017).

# 5 Cultural evolution: how languages become learnable

Not only is language acquired during one's lifetime, but it also evolves over time. Children learn how to walk in more or less the same way today as they did a few centuries ago, but what they learn about the lexicon and the grammar of the same language (e.g. English) is significantly different. Pronunciation may have changed, a lot of new words may have appeared while some others are not used any more, and grammar structures as declination may have disappeared for example. This type of evolution happens too fast to be accounted for by biological evolution, some phenomena even settling in matters of weeks (e.g. invention of new words within a population). Biological evolution has its roots in genetic material being copied and passed on from body to body, whereas cultural evolution takes the form of ideas, words or conceptual structures being passed on from brain to brain. The cultural evolution of language derives from the repeated interactions between individuals of the population using a language, everyone learning from and adapting to their interlocutor. These interactions can be between people of the same generation, designated as horizontal transmission, or different generations, what is called vertical transmission (see Figure 2).

**Cultural evolution can shape languages.** The question is, in the wide landscape of possible configurations and structures, are they all equivalent and only the result of a slow drift of languages? Or like biological evolution, do some of these structures have an advantage, in particular concerning language acquisition? If we look for example at the distribution of vowel systems (De Boer, 2001; Oudeyer, 2018), word order (Ferrer-i-Cancho, 2015), word frequency (Zipf, 1949), or even lexicon size for the domains of color (Kay, Berlin, Maffi, Merrifield, & Cook, 2009) or kinship (Kemp & Regier, 2012), some structures are far more frequent than others in natural languages, which supports the latter hypothesis. Two main pressures exert on language during this evolutionary process: Expressivity, how well a language fits to the context of its usage, and learnability, how well a language can be transmitted and acquired by new learners. A third pressure, closely related to the previous ones, is the facility to reproduce the language in terms of both precision and energy needs, highly dependent on the characteristics of the articulo-auditory system of the agents.

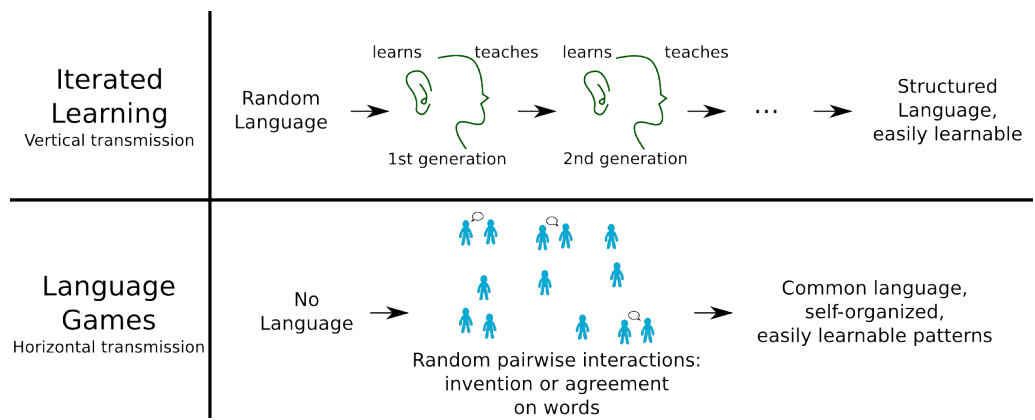

Figure 2: Illustration of the 2 main classes of computational models of language cultural evolution, and how theses processes impact language learnability: Iterated Learning and Language Games

**Language Games and horizontal transmission.** Language Games are a set of computational models in which a population of individuals have pairwise interactions while trying to build or learn a common communication system (Steels, 2001; Loreto, Baronchelli, Mukherjee, Puglisi, & Tria, 2011). For each interaction of a typical Language Game, the corresponding pair of simulated agents — randomly selected from the population — are assigned roles: one is the speaker, uttering a word to refer to a selected meaning or scene, and the other is the hearer, trying to guess what the speaker was referring to. Within a certain number of interactions, that typically depends on the size of the population, all agents agree on a common language and succeed in communicating efficiently. In other words, while the rules underlying interactions remain simple, a communication system can self-organize. For many of the models, convergence towards a shared vocabulary has not only been observed in simulation, but also proven mathematically (De Vylder, 2007; Baronchelli, Felici, Loreto, Caglioti, & Steels, 2006). Moreover, the resulting linguistic structures can show interesting properties, like categories that are well-fitted to both the environment and the sensorimotor system of the users. For example in (De Boer, 2001) and (Oudeyer, 2018), a population of simulated individuals commonly acquire a vowel system. The resulting vowels are always well distributed over the continuous space of possible vowels, and are therefore easily learnable for a new individual that would join the population. Vowels are also selected in a way that minimizes the articulatory energy needed to produce them. A third evolutionary pressure is resistance to noise: because of the non-linearity of the articulatory system, some configurations may be more unstable and sensitive to small variations in the motor commands of the articulatory system. The more stable ones are selected during the evolutionary process. Lending credence to these results, the statistical distribution of the number of vowels over numerous simulations resembles the same distribution in natural languages. Another example is the collective negotiation of names for colors, modeled in different ways: e.g. (Steels & Belpaeme, 2005; Puglisi, Baronchelli, & Loreto, 2008). In particular, the model used in (Baronchelli, Gong, Puglisi, & Loreto, 2010) arrives at a distribution of color categories that are both adapted to the human eye and the frequency of colors in the environment. This model also fits real data, with the average number of color categories produced by the model matching what is observed in the World Color Survey (Kay et al., 2009). Language Games have been used to model many other parts of language, including spatial representation (Spranger, 2012) and grammatical structures (Van Trijp, 2012), and many times the simulated agents are made to interact using real robotic bodies (Spranger, 2012, Steels, 2001).

**Iterated Learning and vertical transmission.** Another type of models, called Iterated Learning, focuses on language transmission between generations (Kirby, Griffiths, & Smith, 2014). Here, simulated agents also interact in pairs but in chains: each of them represents a generation, and only interacts with the previous generation and the next one. A first random model of language is generated and a set of examples of usage of this language are shown to the first generation. Not all possible objects are found in the examples: learning individuals have to generalize from a reduced set only. This is an important part of the models, called the *transmission bottleneck*, which leads to modifications of the language at each generation. The first generation later uses what they have learned about the language to generate a new set of examples that is used to teach the language to the second generation. The generalization process relies on built-in cognitive biases of the agents: they tend to prefer compressible languages.

This process is allowed to proceed for a chosen number of generations, and the language of the final generation is observed. The exact definition of Iterated Learning sums it up quite well: *Iterated learning is the process by which a behavior arises in one individual through induction on the basis of observations of behavior in another individual who acquired that behavior in the same way* (Kirby et al., 2014, pp. 108). Typically, the objects that individuals have to refer to are combinatorial: they have for example a shape and a texture, and all combinations of possible shapes and textures can be found. The final languages range from holistic, with a distinct word for every possible object, to completely structured, with a word for each shape and a word for each texture, the name of the object being a combination of the two. In Kirby, Tamariz, Cornish, and Smith, 2015, it has been shown that with both pressures of expressivity and learnability, structured languages are selected. With only one of the two pressures, they tend to be either holistic or degenerate (with one single word for everything). Even if the starting language is random, the preferred structure is selected and shaped over generations. This illustrates another mechanism of cultural evolution: some patterns are favored and progressively selected because of cognitive biases, and because of these very biases are easier to acquire by new learners having them has well.

Those models do not pretend to describe the full process of language evolution, as they each focus on some specific aspects of language evolution. Therefore they do not represent real language evolution as a whole. However, by studying them we understand that simple mechanisms are enough to observe formation and self-organization of languages. Specific patterns and structures emerge and can be selected, which in turn facilitate language acquisition. This provides a theoretical perspective from which one can interpret the relative ease with which children acquire language.

# 6 Conclusion

Modeling the development and learning of language has inspired researchers in computer science, psychology, and robotics to adopt diverse approaches to the many challenges. We have sought to highlight the main modeling approaches along with the behaviors and empirical data they seek to explain, while also outlining the remaining gaps that remain between these accounts, where future research must be aimed.

For example, cognitive models of cross-situational word learning carried out in the psychology lab typically assume that words and referents are trivially identified and segmented, and that words always appear with their intended referents–assumptions which are often violated in real-world scenes. While more complex developmental robotics models rarely make these assumptions, both cognitive and robotics models are typically only applied to matching human behavior in small-scale learning scenarios, involving short utterances, a few objects at a time, and a total vocabulary of tens of words. In contrast, other studies use mathematical analysis and simulations of learning a full-sized vocabulary, but often make oversimplifying assumptions about the distribution of words, referents, and even the cross-situational learning mechanism, while only attempting to match gross overall human learning rates. Future studies will need to investigate how well robotics models combine with cognitive models to account for both detailed short-term human learning behavior of vocabulary, as well as long-term learning in real-world scenes with full-fledged language and grammatical structures.

Another open dimension of research concerns computational modeling of the discovery of speech as a linguistic tool to communicate with others about referents, and achieve joint tasks. Indeed, most existing computational models (there are few exceptions) have so far relied on cognitive architecture models where language is implicitly assumed to be a system of labels associated to communicative referents. However, for early developing infants, speech sounds (like gestures) are initially part of a rich, unorganized and continuous flow of multimodal information: the special communicative status of these sounds (or gestures) is only progressively discovered. This also highlights the need to develop further computational theories of the ways language development is embedded within the broader picture of sensorimotor, cognitive and social development.